\documentclass[journal]{IEEEtran}

\usepackage{url}

\usepackage{cite}

\usepackage[utf8]{inputenc}

\usepackage[pdftex]{graphicx}
  
\begin{document}

\title{Machine Learning Techniques for MRI Data Processing at Expanding Scale}%

\author{Taro Langner}
	


\maketitle



\begin{abstract}
Imaging sites around the world generate growing amounts of medical scan data with ever more versatile and affordable technology. Large-scale studies acquire MRI for tens of thousands of participants, together with metadata ranging from lifestyle questionnaires to biochemical assays, genetic analyses and more. These large datasets encode substantial information about human health and hold considerable potential for machine learning training and analysis. This chapter examines ongoing large-scale studies and the challenge of distribution shifts between them. Transfer learning for overcoming such shifts is discussed, together with federated learning for safe access to distributed training data securely held at multiple institutions. Finally, representation learning is reviewed as a methodology for encoding embeddings that express abstract relationships in multi-modal input formats.
\end{abstract}

\section{Introduction}\label{sec1}

Deep learning has faced widespread adoption for medical image analysis tasks ever since the breakthrough of convolutional neural networks for image recognition in 2012 \cite{krizhevsky2012imagenet} \cite{dan2012multi}, and especially the publication of the U-Net \cite{ronneberger2015u} for biomedical image segmentation in 2015 \cite{litjens2017survey}. As a technology, it has since also become substantially more accessible due to increased availability of information, abstraction through evolving software frameworks, and ease of access to hardware acceleration with consumer-grade graphics cards or cloud computing resources. Together with open-access repositories for medical imaging data, this has put practitioners, students, and enthusiasts in a position to implement and train deep learning models, within a span of hours, that have the potential to reach human expert performance for a growing number of specific image analysis tasks \cite{litjens2017survey}.

There are several key challenges involved, however, that lead to only a small minority of such projects ever achieving an impact beyond experimental or research settings. For the practitioner, acceptance management and workflow integration can pose the perhaps greatest challenges for applications in medicine. Regulatory constraints and considerations of data governance furthermore impose restrictions on access to specific medical data and can prevent technologically sound approaches from receiving approval for deployment. 
This chapter more closely examines the methodological challenge of generalization across different imaging cohorts and the special case of inference at scale on large, homogeneous imaging datasets such as clinical trials and cohort studies. It also explores how transfer learning can provide benefits from training on comparable large datasets by reusing the acquired information on downstream tasks with potentially more limited access to training data. With such data often being heavily access-restricted, federated learning is furthermore discussed as a way of providing safe access to sensitive data siloed at hospitals and research institutions without infringing on patient rights or privacy concerns. Finally, a brief overview of representation learning is provided for overcoming missing input data and establishing common links between abstract concepts encoded in multi-modal input such as images, text and more.

\section[Large-scale data processing on clinical or epidemiological cohorts]{Large-scale data processing on clinical or \\ epidemiological cohorts}
Large-scale medical imaging studies can examine thousands of volunteers with standardized imaging protocols that are replicated across multiple sites. UK Biobank \cite{sudlow2015uk}, for example, has collected extensive medical data of more than half a million volunteers, including imaging data for a subgroup of 100,000 participants of whom about 70,000 are furthermore planned to undergo follow-up imaging at a later point in time \cite{littlejohns2020uk}.  Similarly, the German National Cohort (NAKO) \cite{bamberg2015whole} aims to acquire MRI of 30,000 volunteers. With time, ever more large-scale MRI datasets are becoming available \cite{lundervold2019overview}. With multiple imaging protocols, a need for quality control, and a plethora of research hypotheses that link the imaging data to extensive metadata ranging from lifestyle questionnaires over biochemical assays to genetic information, the limitations of manual effort for solving these tasks quickly become evident. At these scales, spending just one second of analysis time per image can accumulate to an entire week of full-time work or more. 

Many potentially interesting analyses, such as semantic segmentation of organs, muscle and tissues, require far more than a single second, ranging from minutes to hours or even days when performed with no automation by a trained expert \cite{fischl2002whole}. Whereas a single expert annotator would consequently be required to spend several years or decades on such a task, spreading the work across a wider team poses a challenge to repeatability, with both options incurring significant cost and labor. Whereas these approaches are nonetheless still pursued in the industry, they are often augmented with increasingly powerful semi-automated image analysis techniques\cite{wilman2017characterisation} \cite{linge2018body}. 

Beyond manual and semi-automated analysis, the high degree of standardization in these studies in terms of imaging devices, protocols, and data formatting poses almost ideal conditions for machine learning methodology. Under these controlled conditions, major sources of variability are eliminated, enabling the collection of training data that is directly comparable to those cases that will be encountered by a resulting model when deployed for inference. At the same time, it becomes possible to scale up the quantity of training data, utilizing from dozens \cite{basty2020automated} \cite{langner2020kidney}, over hundreds \cite{liu2021genetic}, \cite{kart2021deep}, and thousands \cite{bai2018automated} \cite{haas2021machine} to tens of thousands \cite{langner2020large} of volumetric scans. A quickly growing body of literature has been dedicated to these tasks that could in all likelihood be covered in an entire book of its own. Rather than attempting an exhaustive review here, only a few examples will be examined in the following.

\subsection{Large-scale applications of machine learning}
A common application of machine learning in this setting is the partial or complete automation of tasks that would otherwise be prohibitive in cost and effort due to sheer scale, even if they could be performed manually at least in principle. But beyond providing what is predominantly an efficiency boost or time savings that merely replace manual effort, machine learning techniques can also be used in this scenario to augment human abilities and perform tasks that are conceptually infeasible for reasons other than simple scalability. One example for this consists of age estimation from brain MRI with neural networks for image-based regression, which not only reached high accuracy but also provided brain-predicted age as a biomarker, with systematic overestimation being associated with premature aging related to injury, degeneration, and mortality \cite{cole2018brain}. Later work also linked this biomarker to genetic information based on genome-wide association studies \cite{jonsson2019brain}. Other work utilized saliency analysis \cite{selvaraju2017grad} for visualizing relevant image features for a given prediction among a larger group of different subjects. This approach has also been combined with image-based regression for properties such as age, highlighting age-related anatomical structures with aggregation across entire subgroups of a given study \cite{langner2020large}, \cite{hepp2021uncertainty}.

\subsection{Quality control of predictions}
Once a model has been deployed for inference on a large-scale study, it can generate thousands or even tens of thousands of predictions. At this scale, anomalies and outlier cases must be expected, along with flawed or entirely failed predictions. For automatically generated segmentation masks, quality control could be performed manually by visual inspection of every single two-dimensional image slice, together with its proposed segmentation mask. In the literature, quality control of this type found nine out of ten images to be error-free, requiring no further corrections \cite{kart2022automated}. However, identifying the remaining failure cases and providing objective quality metrics for the model performance during deployment accordingly represents an important factor in making the generated results reliable as a whole. The paradigm of Machine Learning Operations, or MLOps, combines several best practices for this under the concept of a monitoring phase \cite{makinen2021needs}. The scientific literature has likewise devoted increasing interest to methodological tools for addressing this challenge, providing automated approaches not only for image quality assessment but also for evaluations of machine learning prediction results. For medical image analysis, this typically involves a major overlap with image quality control as a first step, which has already been described in more depth in a previous chapter. In the following, the focus is instead of evaluating the actual predictions themselves.

For semantic segmentation, a common way of identifying failed predictions consists in statistical evaluation of the resulting segmentation masks and underlying image properties. This type of evaluation can also utilize a combination of statistical image features relating to aspects like texture and expected intensities in combination with a machine learning classifier such as random forests trained to distinguish between correct and incorrect segmentations \cite{alba2018automatic}. 
Another approach consists of training a separate convolutional neural network to provide, from an entire given image together with its previously predicted segmentation, an estimate of the Dice score for each predicted class \cite{robinson2018real}.
As an alternative, the concept of Reverse Classification Accuracy (RCA) was proposed to assign a quality metric to each test case by using it, together with its prediction, as a minimal training set for a separate, smaller model referred to as RCA classifier \cite{valindria2017reverse}. This RCA classifier is  evaluated against the original ground training truth data. Its performance metrics on the ground truth data are then used to assign a quality score to the test case itself, under the assumption that only a high-quality prediction would enable for it to learn the segmentation task. For image segmentation, this RCA classifier can be implemented as an atlas segmentation \cite{robinson2019automated}. 

Other approaches use the model itself to generate not only a prediction, but also a measure of its quality with the help of uncertainty quantification. As one such technique, Monte-Carlo dropout \cite{gal2016dropout} can provide a measure of empirical variance by utilizing dropout layers in a model not only during training but also during testing. By keeping the dropout layers active, a random subset of neuron activations will be suppressed, causing multiple repeated predictions to produce slightly different results. This empirical variance can be interpreted as a measure of uncertainty, as high variability for a given prediction may imply low confidence of the model. This approach has also been proposed as a tool for quality control in large-scale MRI segmentation \cite{bard2021automated}. Whereas this methodology models a distribution with an empirical variance by repeated inference, other techniques can treat each prediction directly as a distribution with a corresponding specific, or heteroscedastic, uncertainty of its own \cite{kendall2017uncertainties}\cite{lakshminarayanan2017simple}. This approach has also been utilized to provide automated quality estimates or prediction intervals on large-scale MRI \cite{hepp2021uncertainty} \cite{langner2021uncertainty}.

\subsection{Real-world impact}
These methodologies represent only a small subset of proposed approaches and published work, with ever more sophisticated and powerful methods being developed. Although the literature is speeding ahead of the adoption in the industry and endorsement in the official data catalogs of the underlying large-scale studies, results from machine learning in this space have nonetheless already left an impact on medical research, producing measurements and biomarkers that have formed the basis for further hypotheses and experiments. At the time of writing, UK Biobank in particular has increasingly already admitted into the official data catalog, for sharing with international researchers, a range of derived data fields based on techniques like those discussed in this section \cite{suinesiaputra2018fully} \cite{liu2021genetic} \cite{langner2020kidney}. 

Playing to the strengths of machine learning, these approaches were often able to analyse the entire imaged cohort with a single model for a given endpoint. It would be natural to expect that these models, trained on unusually large and well-curated medical imaging datasets, would now represent an ideal, fully-automated solution for analysis of images from any other study in the future. However, this expectation can typically not be met and these models tend to generalize only to a narrow focus of specific images that closely resemble the original training data. Although they may provide excellent performance on the domain that they were trained for, even minor changes in imaging protocol, device type, or patient demographics may radically change their behavior during inference. The capabilities of common models to generalize can be negatively impacted even when moving between whole-body MRI of studies as superficially similar as UK Biobank and the German National Cohort study \cite{gatidis2023better} due to the phenomenon of distribution shifts, which will be examined more closely in the following section.

\section[Distribution shifts and correction between imaging cohorts]{Distribution shifts and correction between \\ imaging cohorts}

A fundamental assumption of supervised learning is the ability of a machine learning model to learn a given task from a limited set of training data with available ground truth annotations. Once training is complete, the model is expected to generalize and provide reliable predictions for new, previously unseen data when deployed for inference. To avoid simple memorization of the training samples, this ability is typically evaluated on validation or test data that was not included in the training process. However, even excellent evaluation results on validation or test data are no guarantee for robust generalization during deployment. In medical image analysis, it is therefore not uncommon to observe human expert-level performance on one study with a model that nonetheless fails to produce any plausible predictions on images from another dataset, based on subtle differences in image characteristics that may not be evident to the human eye \cite{guan2021domain}.


\subsection{Types of distribution shifts}
Covariate shifts occur when the features associated with a desired label change during deployment in ways not represented among the training data. In MR imaging, this is a common occurrence due to the variable contrast and qualitative nature of image intensities. Variations in imaging protocols, magnetic field strength, coil placement, and imaging device manufacturers can contribute to these shifts. Likewise, changing fields of view, but also medically relevant characteristics of the imaged anatomy such as subject demographics and anomalies like implants or pathologies can play a role \cite{hagiwara2020variability}.

Label shifts refer to changes in the distribution of the property to be predicted in a given domain relative to its distribution among the training data. A typical example consists of changing class distributions, where a model may have been trained on a balanced dataset before being applied to highly imbalanced data.

Concept shift relates to a definitional change of desired labels. In medicine, this can be a common occurrence as diagnoses or even anatomical delineations can vary between experts of the field, with consensus decisions trying to strike a compromise between different subjective interpretations. A model for semantic segmentation of the liver may, for example, fail to meet expectations when the desired behaviour on partial volume, blood vessels, or cysts diverges from the ground truth style provided as training data. Even with just a single operator creating manual reference annotations, variability and style drifts can occur over time \cite{withey2007medical}.

\subsection{The challenges of generalization in CT and MRI}
At first glance, covariate shift in particular could be addressed by aggressive expansion of the training domain and aggregation of extensive datasets capturing the widest-possible variability of data for any given task. Efforts in this direction have enabled substantial progress in the field of natural language processing \cite{devlin2018bert} and also produced models for general-purpose image segmentation \cite{kirillov2023segment}. Despite training on millions of samples, the latter have nonetheless been reported to perform worse on medical image data than models trained for this specific domain\cite{roy2023sam}. To this point, a common curation step for training data in medical image analysis nonetheless consists in the isolation of imaging modalities, with benchmark challenges reporting superior top performances for modality-specific models as compared to those trained on combined data of, for example, both CT and MRI with the goal of being able to accurately segment both image types with a single model \cite{kavur2021chaos}. Ongoing research is continually pushing the boundaries in this area and may well produce more universally viable datasets and approaches, some of which will be examined later on in this chapter. 

For CT imaging in particular, the quantitative nature of Hounsfield units can eliminate one major aspect of variability between images. This has enabled models such as Totalsegmentator \cite{wasserthal2023totalsegmentator} to achieve robust performance, out-of-the-box, on CT data from various sources, leading to widespread sharing of its freely available implementation for segmentation of over one hundred anatomical structures. MR images, in contrast, tend to exhibit greater variability due to the qualitative nature of image intensities and variety of available protocols, field strengths, and scanner settings \cite{hagiwara2020variability}, which makes the training of general-purpose models more challenging. 

When it comes to the reliability of a machine learning model, it would be desirable to provide a formal verification of an expected minimum accuracy across future data, especially from the perspective of acceptance management and regulatory requirements. However, the complex interplay between modern neural network architectures, with potentially millions of trainable parameters, and volumetric images with tens of thousands of voxel intensity values makes it difficult to provide any universal proof for robust generalization under domain shifts. Even though techniques have been proposed for quantifying domain shifts in medical image analysis \cite{stacke2020measuring}, it is typically infeasible to provide any guarantees without applying the model to gather empirical results on the actual data in question. 

\subsection{Correction and training for cross-study analysis}
Whereas distribution shifts between different data sources can affect the performance of a given model, they can also affect medical measurements obtained from source-specific models. Demographic factors such as age, sex, and body weight can have a direct impact on image-derived phenotypes such as measures of organ volumes and body composition, which may moreover be affected by the appearance of the underlying image data. For cross-study analysis, this has prompted the development of data harmonization techniques with model-based approaches that can calibrated one study to the properties of another \cite{gatidis2023better}. Data transformation has been proposed as another technique to make the target domain more similar to the samples encountered in training, which can involve transformations of the feature space utilized by a model, for example by principal component analysis \cite{cheplygina2019not}.

Apart from transforming the data and image-derived measurements, other methodologies aim to improve robustness to domain shifts during training. Here, many common techniques that were designed for improved generalization can also provide immediate benefits for the robustness of a model. This can range from using a suitable pre-processing strategy with regularization and data augmentation in training to techniques like ensembling and test-time augmentation during inference. The highly successful nnU-Net \cite{isensee2021nnu}, for example, resamples images to a common resolution to be more robust against variability in image spacing. It also converts the image intensities of modalities such as MRI to z-scores, making images from different scanners and protocols more comparable. Pre-processing like this can eliminate several aspects of variability in the data before it even reaches the underlying neural network itself. During training, its augmentation strategy with random scaling, elastic deformations and brightness changes can likewise expose the network to samples resembling the results of different scanners or patient demographics. Fundamental techniques such as these, which are primarily intended for improved generalization in the target domain, can often help overcome minor domain shifts when trying to perform the same task on a slightly different domain.

Data synthesis is yet another approach that has been proposed as a tool for overcoming distribution shifts \cite{yi2019generative}. One of its applications is the generation of new samples and augmentation during training, which has been proposed for tasks with imbalanced class labels \cite{shin2018medical}. It has also been proposed for the imputation of missing data or entire additional modalities with cross-domain image generation, for example by performing image-to-image translation with generative adversarial networks \cite{armanious2020medgan} \cite{ben2019cross}. 

With these methodologies, different approaches have been laid out for training techniques that aim to improve generalization or modify the input or output data of a machine learning model to improve its robustness when moving from one domain to another. A common property of the techniques examined here is that the training is entirely constrained to the original domain. A different strategy extends beyond this domain border with the help of transfer learning that utilizes information from one domain to more effectively train a model on another.

\section{Transfer learning}

The availability of training data is often a limiting factor for machine learning in medicine, especially when ground truth annotations for specific tasks are required. One common strategy to address this challenge consists of transfer learning, which can re-purpose knowledge from related upstream tasks to more effectively address a given downstream task. A wide range of techniques have been developed under this concept \cite{zhuang2020comprehensive}, some of which have also been proposed for medical image analysis \cite{van2014transfer}\cite{raghu2019transfusion} and MRI in particular \cite{valverde2021transfer}. 

One application of transfer learning is domain adaptation, where a model is applied to a downstream task of another domain with the goal of overcoming distribution shifts \cite{guan2021domain}. This can involve single-shot and few-shot techniques, in which either one or a small number of labeled training samples are available from the downstream task\cite{wang2020generalizing}. In the extreme case of zero-shot learning, a model is applied to a new task that is posed with no additional downstream training data \cite{palatucci2009zero}. Addressing these challenges tends to require models that incorporate vast amounts of information from upstream tasks with extensive training data, prompting the search for ever more comprehensive foundation models that can provide a starting point for transfer learning approaches.

\subsection{Supervised pre-training on ImageNet}
A common technique for convolutional neural networks in particular has consisted of supervised pre-training on large-scale visual recognition challenge datasets such as ImageNet \cite{kolesnikov2020big} \cite{russakovsky2015imagenet}. Whereas this type of model can be trained without prior knowledge by initializing its trainable parameters, or weights, with a randomized distribution of values, copying the parameters from a model that has already converged on another upstream task can enable potentially faster convergence and improved generalization during inference in the given downstream task. This can occur even when the upstream and downstream tasks are substantially different. As such, supervised pre-training by image classification on millions of images of cars, fruits, and animals can provide advantages even for tasks in medical image analysis \cite{van2014transfer} \cite{shin2016deep}, \cite{raghu2019transfusion}. 

Although models trained on such seemingly unrelated upstream tasks may have never learned to process medical images, the configuration of trainable parameters that enabled their training to converge on another task may still outperform a model initialization that is entirely random. In the case of convolutional neural networks trained on ImageNet, the first network layers have been observed to learn general structural image patterns resembling Gabor filters or color blobs, whereas the final layers tend to be more task-specific \cite{yosinski2014transferable}. Learning to extract such general-purpose features may accordingly provide a better starting point for further optimization on a new downstream task than a random weight initialization. On medical image data, these features have been reported to evolve into less generic patterns during continued training or never arise at all when training from scratch \cite{raghu2019transfusion}. Structural benefits can arise even from pre-training on an upstream task for which labels themselves have been generated at random \cite{maennel2020neural}.

For the model to be initialized with pre-trained weights, differences between the upstream and downstream task can require changes to the neural network architecture, so that not all pre-trained weights can be copied directly. As an example, a convolutional neural network trained on the ImageNet \cite{russakovsky2015imagenet} visual recognition challenge to predict logits for 1000 different classes may require for its last fully-connected layer to be replaced if the new downstream task only requires two class labels \cite{raghu2019transfusion}. Similarly, its first convolutional layer with three channels for RGB input may have to be pruned or replaced if the goal is to train on a downstream task with only one input channel. When replacing model components in this manner, the affected weights can either be reused in part of replaced entirely with random initialization.

Due to the benefits of supervised pre-training, various approaches have been developed to pose a given downstream task to resemble the original upstream task as closely as possible. For medical image analysis, this can involve utilizing a model in a way that closely resembles the processing of natural RGB images as encountered in ImageNet \cite{russakovsky2015imagenet}. Conceptually, posing the downstream task in this way aims to keep the distribution shift to the upstream task as small as possible. One common step for this consists in pre-processing to scale or limit the image intensities to a specific range. Often more challenging is the disparity in the image shape, which for medical data can often be volumetric with potentially more or fewer than just three RGB color channels. Approaches for bridging this gap can involve 2.5D approaches that combine different slices of a volumetric images into a single, two-dimensional image with three channels. These slices can be orthogonal patches that provide different views on the same anatomical point in multi-view fusion \cite{roth2015improving}, three consecutive slices of the same imaging plane \cite{isensee2021nnu}, or combinations of different imaging modalities\cite{hesamian2019deep}.

Once a model has been successfully initialized with pre-trained weights, different strategies have been proposed for subsequent use on the downstream task\cite{shin2016deep}. One approach consists of merely applying the model as a feature extractor without performing any further training. The resulting "off-the-shelf" features can then be utilized by other means and potentially be combined with non-imaging data as proposed for radiomics applications \cite{avanzo2020machine}.
Other approaches limit the training to only a subset of model weights. By keeping the remainder of the model unchanged, or "frozen", a complex model can be trained to adapt to a more limited pool of downstream data without greater risk of overfitting. Variations of this approach can range from freezing all but the final layer for fine-tuning to using different learning rates that inhibit the adaptation of some layers\cite{shin2016deep}. Perhaps most commonly, the entire model can also be trained with no further restrictions, so that the initialization with pre-trained weights merely serves as a deliberate starting point. Other work has more recently started to utilize weights of ever more comprehensive upstream tasks involving multiple modalities, such as images and text, which will be examined more closely later on in this chapter under the aspect of representation learning.

In modern software frameworks, supervised pre-training can also be performed implicitly by offering a range of existing model weights as an initialization option rather than requiring for any computational resources to be spent on actually training on an upstream task. Whereas these options can cover a wide range of popular benchmark challenge datasets that received extensive attention in research, they have historically favoured neural networks architectures for natural, two-dimensional image data with colour channels. In turn, more specialised architectures with volumetric filter kernels that have achieved substantial success for voxel-based image data in medicine are not as commonly represented. Ongoing research efforts are aiming to change this by collecting more extensive volumetric medical imaging data for the purpose of pre-training suitable foundation models for medical imaging data \cite{chen2019med3d} \cite{vistaGitHub} \cite{ye2023uniseg}.

\subsection{Positive and negative transfer}
Positive transfer is said to occur when transfer learning benefits the downstream task. However, it is worth noting that transfer learning is not always a required prerequisite for good performance. There is no universal guarantee for transfer learning to provide performance benefits, and it can indeed cause negative transfer to occur in which the downstream task is solved worse than it would have been with a random weight initialization \cite{raghu2019transfusion}. At the time of writing, volumetric, convolutional architectures such as nnU-Net\cite{isensee2021nnu} have established state-of-the-art results in a wide range of medical benchmark challenges even without being able to reap the full benefits of pre-training. Likewise, even transformer models such as UNETR reported no benefit from using pre-trained weights for medical data \cite{hatamizadeh2022unetr}, despite the underlying Vision Transformer (ViT) \cite{dosovitskiy2020image} architecture having been proposed with heavy reliance on pre-training for natural images. 

Beyond transfer learning, multiple related methodologies have been proposed that do not strictly focus on one given target domain. In multi-task learning, for example, one model is trained to concurrently perform several tasks in parallel \cite{zhang2021survey}. This can involve training the same model on different sub-tasks and potentially using different, task-specific, output layers, whereas the remaining model weights are shared between tasks. The model is thereby required to learn features that are relevant to all included tasks, which can have a regularization effect that has been reported as beneficial especially at smaller sample sizes \cite{cheplygina2019not}.

Many more techniques have been proposed in this context, which can help to overcome limited training data in the target domain \cite{valverde2021transfer}. Beyond moving information between different domains, other machine learning methodologies have been proposed that can utilize training data from several sources. As this type of aggregation can pose considerable challenges in the field of medicine, the next section provides an overview of how federated learning can achieve this without any requirement for transmitting, pooling, or otherwise exposing any sensitive data directly.

\section{Federated learning}

Many institutions around the world have exclusive access to medical imaging data, such as hospitals, research organizations, and companies. Despite the wealth of contained information, sharing this data for research is largely restricted by regulatory constraints and concerns relating to data privacy and security. These challenges of data governance mean that only a small minority of the images acquired by thousands of MR devices around the world in daily use are accessible for analysis with machine learning \cite{rieke2020future}. 

Federated learning is a technique with the potential for bridging this gap \cite{mcmahan2017communication} \cite{xu2021federated} \cite{nguyen2022federated}. It allows for the distributed training of models without pooling or exposing the underlying images or records, so that the respective owners can retain full data sovereignty. Instead of transmitting the data itself, this methodology can share the parameters of a global consensus model that is refined locally and aggregated with different compute plans and infrastructure topologies to optimize a global loss function\cite{konevcny2016federated}.

In each learning round, the participating entities of the federation can perform one or more iterative optimization steps of the global loss function by training on the locally available data and applying the resulting parameter updates to their local copy of the model. To conclude the round, the aggregated parameter updates are then shared with the other participants. When a high number of local optimization steps is performed as part of a given round, the overarching training process as a whole tends to increasingly diverge from a conventional approach of training on pooled data and special care must be taken to avoid potential performance penalties. In turn, sharing more granular updates can pose a risk of exposing more directly identifiable elements that could be vulnerable to exploitation or information leakage \cite{wang2019beyond}.

The participating entities can be arranged as nodes into different topologies, such as decentralized architectures in which they communicate directly with each other and the overarching model is updated in parallel or centralized designs in which all nodes are connected to one aggregation server. In the latter, the participating institutions do not necessarily have to be aware of each other and can  remain anonymous or be blinded to one another. Hybrid approaches can furthermore combine both of these structures in hierarchical architectures that can be composed of nested sub-federations. The compute plan determines how the global model is updated based on the local processing results. For example, the update can be performed on a peer-to-peer basis, over the aggregation server or in sequential order between nodes \cite{rieke2020future} \cite{kairouz2021advances}. In practice, the latter is typically avoided due to the risk of catastrophic forgetting, where a model may no longer be able to correctly process training samples that only occurred early on in the training process \cite{french1999catastrophic}.

Semantic segmentation of brain MRI has received special attention for applications of federal learning, with early work reporting competitive results for brain lesion segmentation when simulating different institutions collaborating over a centralized aggregation server \cite{sheller2019multi}. Another approach proposed decentralized, peer-to-peer learning for whole brain segmentation on data of different institutions, enabling shared training on ground truth segmentations that may require up to one week of manual work by expert annotators per case \cite{roy2019braintorrent}. Other work proposed decentralized federated learning as a privacy-preserving approach for combining training data from fMRI of the brain and explored domain adaptation techniques for overcoming domain shifts between different sites \cite{li2020multi}. Further work explored federated learning for MRI reconstruction \cite{guo2021multi} \cite{elmas2022federated} and synthesis \cite{dalmaz2022one}. 

Ongoing work is increasing the scope of comparable initiatives, with the goal of aggregating ever more data for training of machine learning models\cite{rieke2020future}. Beyond segmentation, classification, and reconstruction, there is increasing interest in also linking the underlying images to multi-modal metadata such as textual descriptions and reports that could be explored with more recent methodologies for representation learning. 

\section{Representation learning}

One of the primary reasons for the popularity of deep learning has been the ability of neural networks to learn relevant features for a given task from data through mathematical optimization \cite{bengio2013representation}. In medical image analysis, these automatically learned features have been found to not only outperform hand-crafted features in a variety of downstream tasks, but also to hold potential as generic "off-the-shelf" representations for image data \cite{shin2016deep}. Research has since been exploring further ways to form, manipulate, and utilize these feature representations for various applications including MR image processing.

Among these applications have been attempts to overcome domain shifts by finding representations of MR image features that are invariant to the exact type of scanner and protocol that they were acquired with. Domain-adversarial networks have been proposed as one such approach that can form domain-invariant feature representations by using an adversarial domain discriminator\cite{kamnitsas2017unsupervised}. In this adversarial setting, this discriminator is trained to predict the given domain from intermediate feature representations of a given network. The training is successfully concluded when these feature representations no longer provide any indication to the discriminator as to which input domain they were extracted from. 

Related work similarly proposed strategies for disentangling feature representations into factors that could differentiate between characteristics of the imaging modality, such as CT or MRI, and an underlying spatial and anatomical structure that could then be processed independently of which modality it originates from \cite{chartsias2019disentangled}.  This concept has also been leveraged to process multi-modal MRI in which one or more modalities are missing by embedding the input modalities into a shared modality-invariant latent space \cite{havaei2016hemis} \cite{zhou2021latent}. This not only enabled segmentation with one or more missing input modalities, but could also be used in a generative strategy, for example to create images with synthetic lesions \cite{chartsias2017multimodal}.

\subsection{Language–Image pre-training}
The concept of multi-modal input has received increasing attention and can be taken beyond merely combining different types of images towards concurrent processing of entirely different formats such as images, video, and text \cite{baltruvsaitis2018multimodal}. Following earlier work that examined X-ray images paired with text data \cite{huang2021gloria} \cite{zhang2022contrastive}, Contrastive Language–Image Pre-training (CLIP) \cite{radford2021learning} was a proposed and received substantial attention as a method that could represent both text and images in a multi-modal embedding space. This approach can be pretrained on batches of images, each of which is paired with one corresponding text. The contrastive optimization aims for the similarity between the image embedding as created by an image encoder and the text embedding as created by a text encoder to be maximized for each pair, whereas the similarity between incorrect pairings is minimized. Once pre-training on 400 million input pairs is concluded, the model has thereby been trained to relate visual and textual concepts to each other by converting both input formats into a similar feature embedding\cite{radford2021learning}.

The ability of models like CLIP to learn representations that link text and images together has since been applied for a growing range of applications outside of medical image analysis. One such example consists of image classification in a zero-shot setting, where no training is conducted on the target domain. This task was posed by creating one textual caption for each potential class and comparing its text embedding to the image embedding of a given input image. The class that scored the highest similarity between its embedded caption and the embedded image was then assigned, with an accuracy rivaling specialized architectures trained for the specific task \cite{radford2021learning}. In later work, CLIP-guided synthesis has furthermore been used to control the process of image generation with textual prompts in methods such as Stable Diffusion \cite{rombach2022high}. 

Whereas many of these approaches examined non-medical settings, similar techniques have since been proposed for medical image analysis as well. For organ segmentation and tumor detection in CT, a CLIP-driven universal model was proposed that trained on data of several pooled benchmark challenge datasets with mutually inconsistent labels and could isolate specific targets based on textual prompts \cite{liu2023clip}. This approach was subsequently expanded to CT, MRI and PET data in a continued push towards a universal segmentation model \cite{ye2023uniseg}.

\subsection{Universal segmentation models}
Another effort towards a universal segmentation model has been building on the Segment Anything Model (SAM) \cite{kirillov2023segment}. This approach represents two-dimensional images with embeddings that are then combined with a prompt that can consist of either an input mask, points, boxes or free text to produce multiple output segmentation masks. Whereas the original system was trained on 11 million diverse natural images and its performance on medical data was reported with mixed results \cite{roy2023sam}, more recent work has proposed a closely related approach for medical images that included over one million paired images and segmentations from 15 modalities in training \cite{ma2023segment}. Ongoing work is continually widening the scope of inputs and can benefit from ever more data being available from research organizations and benchmark challenges that has been released for shared use.

\section{Conclusion}
Beyond just automating tasks that would have otherwise required prohibitive human effort, machine learning techniques have added a range of tools for researchers to extract information from large-scale MRI that would not otherwise be accessible. From extracting automated measurements at scale in ongoing studies towards providing new image-derived biomarkers, the examined approaches have already been able to make a tangible contribution to medical research. By learning more universal and sophisticated feature representations, recent machine learning methodologies are becoming increasingly powerful in extracting underlying concepts of human knowledge from diverse types of data and relating them to each other. In the search of universally viable embeddings, these systems can go beyond memorization of specific shapes and patterns towards progressively more abstract relationships to capture information that, while difficult to formalize or encode explicitly, is of essential interest and intuitive use to a human observer, regardless of the particular medium or format at hand. As these techniques are rapidly evolving, they will form a growing set of tools that research can utilize to probe and explore the ever growing quantity and depth of medical imaging data. 


\bibliographystyle{ieeetr}

\bibliography{references}

\end{document}